# MPI: Multi-receptive and Parallel Integration for Salient Object Detection

Han Sun [1,2], Jun Cen [1,2,*], Ningzhong Liu[1,2], Dong Liang[1,2], Huiyu Zhou[3]

[1] College of Computer Science and Technology, Nanjing University of Aeronautics and Astronautics, Nanjing, China
[2] MIIT Key Laboratory of Pattern Analysis and Machine Intelligence, Nanjing, China
[3] School of Informatics, University of Leicester, Leicester LE1 7RH, U.K.
[*] Corresponding author: Jun Cen (nuaacenjun@126.com)

**Abstract:** The semantic representation of deep features is essential for image context understanding, and effective fusion of features with different semantic representations can significantly improve the model's performance on salient object detection. In this paper, a novel method called MPI is proposed for salient object detection. Firstly, a multi-receptive enhancement module (MRE) is designed to effectively expand the receptive fields of features from different layers and generate features with different receptive fields. MRE can enhance the semantic representation and improve the model's perception of the image context, which enables the model to locate the salient object accurately. Secondly, in order to reduce the reuse of redundant information in the complex top-down fusion method and weaken the differences between semantic features, a relatively simple but effective parallel fusion strategy (PFS) is proposed. It allows multi-scale features to better interact with each other, thus improving the overall performance of the model. Experimental results on multiple datasets demonstrate that the proposed method outperforms state-of-the-art methods under different evaluation metrics.

## 1. Introduction

Salient object detection (SOD) aims to identify the most visually distinctive objects or regions that draw human visual attention in natural scenes. It is widely used in a variety of Computer Vision-related tasks. Such as semantic segmentation, image retrieval, visual tracking, and objection detection [1]. Early SOD models are proposed based on cognitive studies of the human visual attention mechanism, color, texture, and global contrast of input images are extracted as discriminatory features, such as SR [2] and RC [3]. However, these hand-crafted features can hardly capture high-level semantic information, and the performance of these models is limited, especially when the contrast between the foreground object and background region is low [4]. With the great success of convolutional neural networks (CNN) [5] in image classification, deep learning has brought a revolution in salient object detection.

Recently, many salient object detection models based on CNN have been proposed with improved performance and most of them use an encoder-decoder structure to extract multi-scale features [6, 7, 8]. Experiments have demonstrated that fusing features at different scales can enhance the expression of feature maps and improve the performance of the SOD models [9]. However, features from different levels have large differences. Lower-level features have less semantics and more details which help to distinguish object boundaries. And deeper-level features have more semantics which is important for region location. Therefore, the effective integration of multi-scale features allows features from different layers to interact with each other, thus improving the performance of the network. The receptive field in a SOD network can be represented as an area on the input image which corresponds to a point in the output feature map. The essential reason for multi-scale features to have semantic differences is that different layers have different receptive fields. A larger receptive field can produce more semantics and higher-level features. Meanwhile, feature maps with smaller receptive fields have rich details. Usually, the convolutional kernel size is closely related to the receptive field, the larger the convolutional kernel is, the larger the receptive field is, but the calculations are also increased significantly. Efficient integration of features with different receptive fields while reducing the computations is important for designing an efficient SOD model.

To address the above problems, we propose a novel salient object detection model named MPI. In order to expand the receptive field and generate feature maps with different receptive fields and semantics, a dilated convolutional group with different dilation rates is applied to each side-output separately. At the same time, the kernel size is constant so as to obtain multiple semantic features without increased computations. The proposed method fuse outputs from adjacent layers progressively instead of a top-to-down way to integrate multi-scale features because the semantics in adjacent layers are similar. In this way, the model can reduce the introduction of redundant information and allow multi-scale features to better complement each other. The main contributions of this paper can be summarized as follows:

1. A multi-receptive enhancement module (MRE) is proposed, which enriches the receptive fields of deep features and enhances the semantic representation.

2. A simple but effective parallel fusion strategy (PFS) is designed, which prioritizes the integration of adjacent layers to reduce the introduction of redundant information and allows features from different layers to better complement each other.

3. Extensive experiments demonstrate that the proposed method achieves the start-of-the-art performance on five datasets in terms of six metrics, which proves the effectiveness and superiority of the proposed model.

## 2. Related work

Before CNN is introduced to SOD, early models require extracting hand-crafted low-level features such as color, central prior, and contrast before using a classification model to obtain a saliency map [10]. With the CNN-based models being proposed, the pixel-wise prediction has taken the performance of SOD models to a new level, and the extraction and fusion of multi-scale features have received extensive attention [11]. Recent SOD methods can be broadly classified into three main categories in terms of feature extraction and fusion architectures, including U-Net [12] style, HED [13] style, and U-Net+HED style.

**U-Net style** [12]. U-Net is first proposed in 2015 for biomedical image segmentation and is one of the first algorithms which use a fully convolutional network for semantic segmentation. U-Net extracts multi-scale semantic features using multiple down-sampling and progressively fuses feature maps from deep to shallow using up-sampling. Then many U-Net-based SOD models are proposed to fuse features from different layers [14]. Luo et al. [14] propose a simplified convolutional neural network which combines local and global information through a multi-resolution 4×5 grid structure. This model enables near real-time, high performance saliency detection. Based on U-Net, [15] proposes a bi-directional structure to pass messages between multi-level features, and a gate function is exploited to control the message passing rate. Compared to U-Net, it further facilitates the fusion of multi-scale features.

**HED style** [13]. Holistically-Nested Edge Detection (HED) is first developed to detect edges. It converts pixel-wise edge classification into image-wise prediction through multi-scale feature learning and holistic image training. Afterward, many HED-like networks have been proposed to detect salient objects. [16] designs an accurate and compact deep SOD network, it first employs residual learning to learn side-output residual features for saliency refinement, then the reverse attention is used to guide such side-output residual learning in a top-down manner. This method has advantages over state-of-the-art methods in terms of simplicity, efficiency (45 FPS), and model size (81 MB). Hou Q et al. [17] find that the structure of HED is not suitable for salient object detection, although it can fuse features from deep and shallow layers well. Therefore, they propose to better fuse multi-scale features by adding short connections on the skip-layers from top to down.

**U-Net + HED style.** The U-Net+HED mechanism combines the advantages of U-Net and HED structures. It not only uses an encoder-decoder structure to generate saliency maps but also adds supervisions at the intermediate output layers to train a multi-scale prediction model. By using a linear combination of multi-layer loss, it can achieve local and global optimality. The final saliency map is a combination of multi-layer predictions. Zhang X et al. [18] proposes an attention-oriented SOD method that can selectively fuse multi-level contextual information progressively. Through multi-path recursive connections, global semantics from the top convolutional layers are passed to the shallow layers. Liu N et al. [19] proposes a novel saliency detection algorithm called PiCA-Net that can selectively focus on informative context locations for each pixel. PiCA-Net contains global and local networks to focus on global and local contexts respectively, and they facilitate the learning of global contrast and smoothness. Many other high-performance models fall into this category [8, 9, 11, 20].

Although the performance of recent models is getting better, problems still remain in salient object detection. First, the multi-receptive representation of features is still inadequate. Expanding the receptive field and making features from the same layer have different receptive fields can make the integrated features more discriminating and more useful for network judgment. Second, multi-scale feature fusion methods are not the more complex, the better. Although the more complex the integration strategy, the better the feature fusion, some of the features may be used repeatedly, leading to more redundant information and increased computation, which may be detrimental to the inference of saliency maps. Based on the above problems, this paper proposes a novel method to efficiently extract and fuse multi-scale features and infer saliency maps accurately.

### 3. Proposed Method

This section first describes the proposed multi-receptive enhancement module (MRE), then explains the Parallel fusion strategy (PFS) in detail, and finally introduces the loss function. MRE can generate multi-receptive features and enhance semantic representation. FPS makes features interact better by fusing multi-scale features in a parallel way. The proposed model uses ResNet-50 [21] as the backbone network to encode semantic features. ResNet-50 is a deep feature extraction network proposed by He et al. in 2016. It consists of five encoding stages, the outputs of each stage represent the semantic features of the input image learned in that stage, and each stage outputs semantic features at half the scale of the previous stage [21]. The overall structure of the network is shown in Fig.1.

#### 3.1. Multi-receptive enhancement (MRE)

Multi-scale feature fusion can complement details to deep semantic features and provide semantics to low-level features. The essential reason for the large variability of multi-scale features is the different receptive fields of the convolutional layer. A large receptive field allows features to contain more semantics, while a small receptive field retains more details [20]. Expanding the receptive field and integrating features with different receptive fields can enhance the semantic representation of feature maps and improve the inference ability of the model.

The five side outputs of ResNet-50 can be denoted as $\{O_i \mid i = 1,2,3,4,5\}$, considering that the receptive field of $O_1$ is too small, the model only uses the output of the second to fifth stages, as shown in Fig.1. Some feature maps from $\{O_i \mid i = 2,3,4,5\}$ are shown in Fig.2.

In MRE, it first converts $\{O_i \mid i = 2,3,4,5\}$ into 128 channels by a 1 x 1 convolution. Unifying the number of feature channels not only reduces the calculations but also is convenient for feature integration. Then MRE uses three separate dilated convolutions with different dilation rates for the same input. In order to keep the computation from incr-easing, the size of the three convolutional kernels is fixed to 3 x 3. It is worth noting that for $\{O_i \mid i = 2,3,4\}$ the dilation rates are 1, 2, and 3, and for $O_5$, the dilation rates are 1, 3, and 5, respectively. Dilation 1 means the kernel works as a standard 3 x 3 kernel. Dilation 2 makes the 3 x 3 kernel sample every 2 pixels. Therefore, the original 3 x 3 kernel is actually expanded to 5 x 5, and the receptive field has also increased significantly. Similarly, dilation 3 and dilation 5 make the kernel expanded to 7 x 7 and 11 x 11, respectively. As the dilation rate increases, the receptive field becomes larger. After independent dilation convolutions, the outputs which have different receptive fields are added by element-wise addition. A 3 x 3 convolution is followed to fuse the features, as shown in Fig.3. Fig.4 shows the original features from $O_5$ and the fusion of results after dilated convolutions with dilation rates 1, 3, and 5, respectively. The features that have been processed by dilated convolutions with different dilation rates have semantics which is more accurate and closer to ground-truth than the original deep features.

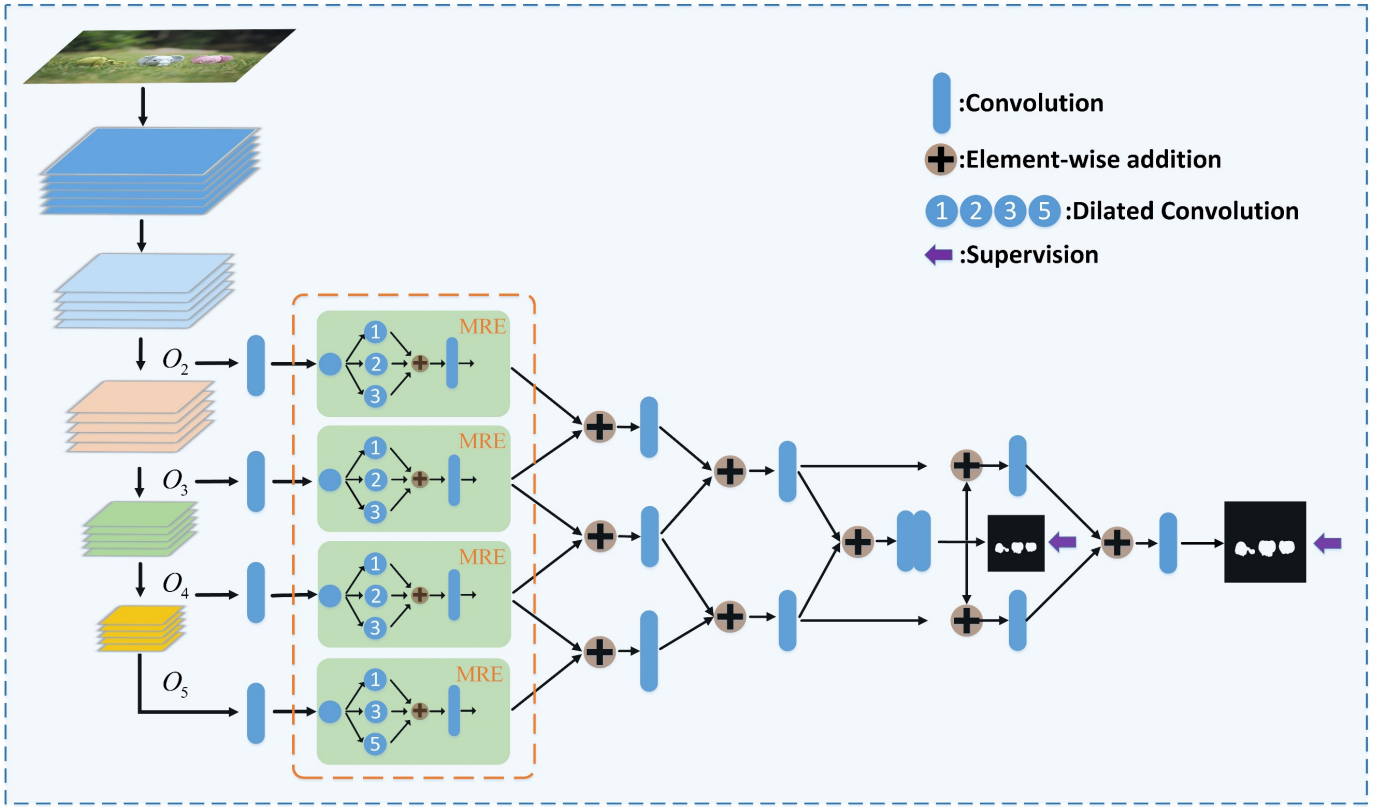

***Fig.1.*** *The overall architecture of the proposed model. Multiple dilated convolution groups with different dilation rates are used to generate features with rich receptive fields. A simple but effective parallel fusion strategy can reduce the introduction of redundant information and allow multi-scale features to interact better.*

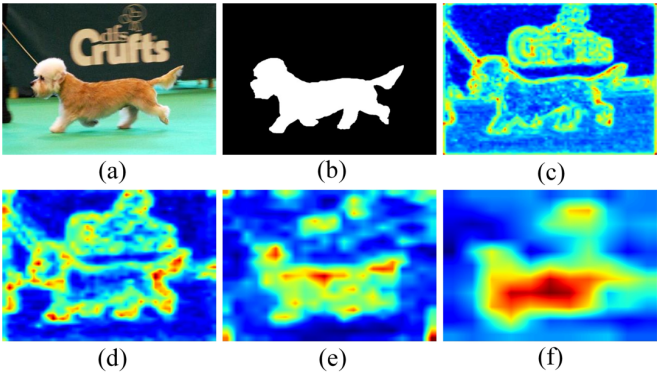

***Fig.2.*** *feature map from $O_2$ (c), $O_3$ (d), $O_4$ (e), $O_5$ (f), respectively. (a) and (b) are image and ground-truth, respectively.*

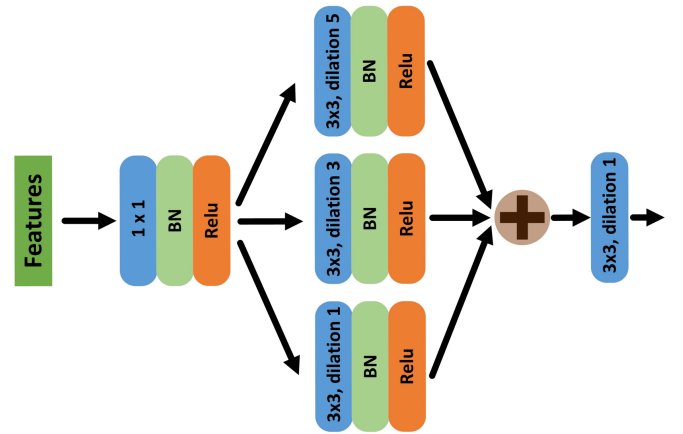

***Fig.3.*** *The MRE generates features with multiple receptive fields based on the features from each side.*

### 3.2. Parallel fusion strategy (PFS)

Recently, many algorithms have been shown to fuse multi-layer and multi-scale features to effectively enhance the feature representation[4, 7, 9, 11, 20]. The performance of the model is significantly improved by the complementarity of deep semantics and details. Nowadays, feature fusion is mostly a top to down approach which passes the semantics from deep layers to shallow layers, such as the very popular U-Net and U-Net + HED architectures, as shown in Fig.5. With the deepening of research, feature fusion is becoming more and more sophisticated, many complicated top-down fusion methods have been proposed, but the complex inter-layer connections lead to an increase in computations. However, the strategy of feature fusion should not be as complex as possible. Often, the intricate connections will cause some features to be used repeatedly, and the repeated use of redundant features may even have a negative impact on the reasoning of the network. In addition, deep semantics are progressively diluted in the top-down transmission process.

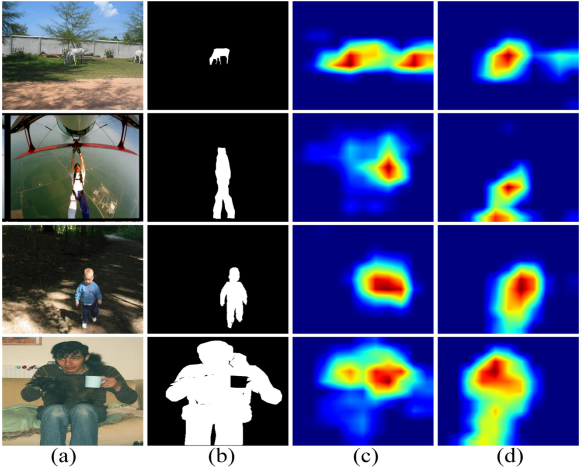

**Fig.4.** *Semantic features. (a) image, (b) ground-truth, (c) features from the fifth side, (d) fusion of results after dilated convolutions with dilation rates 1, 3, and 5, respectively.*

Therefore, we propose PFS to integrate multi-scale and multi-layer features, as Fig.5 (d) shows. Although the structure of PFS is simple, it can effectively avoid the introduction of redundant information and enhance the feature representation by fusing local features to generate an overall feature representation. Considering that the essential reason for the differences in multi-scale features is that the receptive fields of the convolutional layers are different. The closer the receptive fields are, the more similar the semantics of the resulting features are. Hence, the model should give priority to fusing features with similar receptive fields. First,

PFS locally fuses features from adjacent layers that has small differences in receptive fields, and then iteratively integrate the features of the adjacent layers to produce the overall feature representation. By parallel fusion, the two features with small differences are selected each time, which can effectively reduce the introduction of redundant information. The whole process could be shown as follows.

$$f'_k = G(f_i + f_{i+1}), i = 2, \ldots, n-1. \quad (1)$$

Where $f'_k$ is the locally fused features, $f_i$ and $f_{i+1}$ are the selected features from the adjacent layers, and G denotes the Convolutions.

In addition, the model has two predictions and corresponding supervisions, as Fig.2 shows. The results of the first prediction were used as feedback to fuse with the two previous feature maps, and then the two processed features were integrated to predict the final saliency maps. Compared with the first round of prediction, the second round corrects some errors, which makes the results more accurate.

### 3.3. Loss Function

Binary cross-entropy loss (BCE) is widely used in saliency detection models [9]. BCE supervises the model by calculating the classification loss pixel by pixel, and the importance of all pixels is equal. In the actual reasoning process, the error-prone regions should be focused, such as boundary pixels, and increase the penalty for boundary pixels, which will improve the model's performance on object boundary judgments. Therefore, when calculating the loss, we set different weights for different pixels. If the pixel is on the object boundary, then it should be given greater weight, noted as $\alpha$, which is shown in (2) [20].

$$\alpha = 1 + sigmoid(E) \quad (2)$$

Where $E$ denotes the single-channel edge map which is generated by the corresponding saliency mask, as shown in Fig.6.

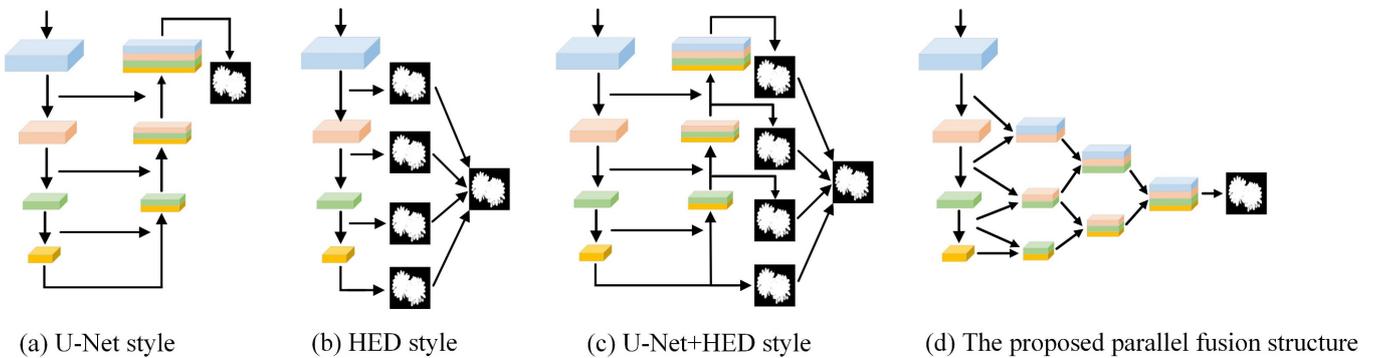

(a) U-Net style   (b) HED style   (c) U-Net+HED style   (d) The proposed parallel fusion structure

**Fig.5.** *Different architectures to fuse multi-scale features. (a) and (c) are top-down approaches, (b) uses multi-level supervisions. (d) is the parallel fusion structure.*

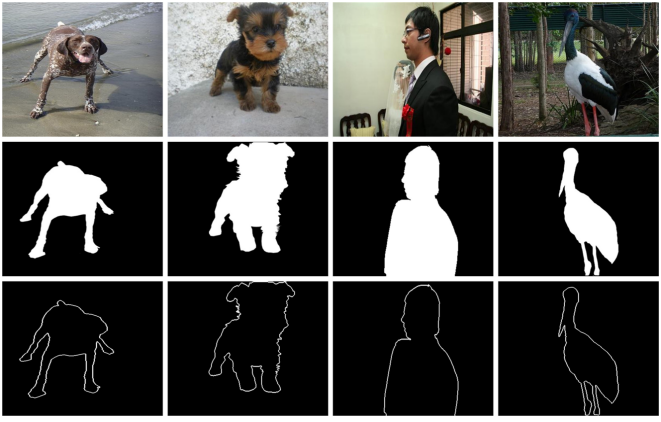

*Fig.6. Some examples of saliency masks and corresponding edge maps. The first row is the image, the second row is the saliency mask, and the third row is the edge map.*

The weighted binary cross-entropy loss (WBCE) function can be calculated as (3) [9].

$$l_{wbce} = -\frac{\sum_{n \in N}(1+\alpha_n)\sum_{l=0}^{1} I(g_n = l) \log P(p_n = l|\varphi)}{\sum_{m \in N} \alpha_m} \quad (3)$$

Where $N = W * H$, and it denotes all pixels of the image, $I(.)$ is the indicator function. The symbol $l \in \{0,1\}$ indicates two kinds of labels. $\alpha_n$ denotes the weight of pixel $n$. $g_n$ is the corresponding ground truth, and $p_n$ is the prediction. $\varphi$ represents the learned parameters.

WBCE loss is able to assign different weights to pixels based on local spatial information but lacks global structural information. Therefore, IoU loss is also used to make the model focus on structural similarity. IoU loss is widely used in image segmentation, it aims to optimize the global structure rather than focusing on individual pixels. Similarly, we use weighted IoU (WIoU) loss to make the model focus more on difficult regions. It can be shown in (4) [9].

$$l_{wIoU} = 1 - \frac{\sum_{n \in N}(g_n * p_n)*(1+\alpha_n)}{\sum_{m \in N}(g_m + p_m - g_m * p_m)*(1+\alpha_m)} \quad (4)$$

$l_{wbce}$ and $l_{wIoU}$ are added as the loss function of each prediction, as (5) shows. While generating the saliency map, the proposed method adds supervision on both intermediate and feedback predictions, and the loss function for the entire network is a linear combination of the two, as shown in (6).

$$l_t = l_{wbce} + l_{wIoU} \quad (5)$$

$$L = \beta * l_t^1 + (1-\beta) * l_t^2 \quad (6)$$

Where $\beta$ is a hyperparameter which is set to 0.5.

We train the same model with different loss functions separately and evaluate it on the ECSSD dataset. The results are shown in Table 1. It can be seen that the performance of the model is close when using BCE or IoU alone while using BCE and IoU together results in a significant improvement. After giving greater weights to the edge pixels based on the combination of BCE and IoU, the model's performance further improved as shown in the last row of Table 1.

**Table 1** Performance of the proposed model when using different loss functions. The best results are bolded.

|  | ECSSD | | | |
| --- | --- | --- | --- | --- |
|  | mF | MAE | $S_\alpha$ | $E_\varepsilon$ |
| BCE | .924 | .035 | .919 | .926 |
| IoU | .925 | .034 | .915 | .920 |
| BCE+IoU | .930 | .033 | .922 | .924 |
| Weighted (BCE+IoU) | **.933** | **.032** | **.925** | **.930** |

## 4. Experiments And Analysis

### 4.1. Implementation

The ResNet-50 [21] pre-trained on ImageNet is used as the backbone to obtain deep features, and the network is trained end to end on DUTS [22] dataset. The proposed algorithm is implemented in PyTorch1.2.0 with cuda9.2, and a TITAN XP GPU is used for acceleration with 12GB memory. The learning rates of backbone and branch are set to 0.002 and 0.02, respectively. The weights of convolutional layers in the branch are initialized by Kaiming-Normal, and the warm-up strategy and linear decay are used to adjust the learning rate. The momentum is set to 0.9 and weight decay is set to 5e-4. During training, the batch size and epoch are set to 16 and 36, respectively. The saliency maps of the two predictions are directly added as the final result without any post-processing.

### 4.2. Datasets and Evaluation Metrics

Five widely used datasets are used to test our model, including ECSSD [23], PASCAL-S [24], DUT-OMRON [25], HKUIS [26], and DUTS [22].

The full name of ECSSD is Extended Complex Scene Saliency Dataset. To represent the situations that natural images generally fall into, Yan et al. extend the Complex Scene Saliency Dataset (CSSD) [27] to a larger dataset (ECSSD) with 1000 images. It includes many semantically meaningful but structurally complex images for evaluation. All the images are acquired from the internet, and 5 persons were asked to produce the ground truth masks.

The PASCAL-S dataset is built on the validation set of the PASCAL VOC 2010 segmentation challenge [28], and it contains 850 natural images. When making the masks, the observers are asked to select the salient objects by clicking. There is no time limit and no constraint on the number of objects that each person can select.

DUT-OMRON is a very challenging SOD dataset proposed by Yang et al. [25]. It contains 5168 high-quality images, but most images have multiple salient objects and complex backgrounds.

HKUIS contains 4447 high-quality images, and all the images have high-precision labels. Images in HKUIS are often equipped with multiple objects and many objects close to the image boundary. Hence, it brings great challenges to SOD methods.

DUTS contains 15572 images, including 10553 images for training and 5019 images for testing, which is the largest saliency object detection dataset. All the training images are selected from the ImageNet DET [29] training/val sets and most of the test images are collected from the ImageNet DET test set. The pixel-level ground truths are manually annotated by 50 subjects.

Mean F-measure (*mF*), mean absolute error (MAE) [30], S-measure [31], and E-measure [32] are calculated to evaluate the proposed method. *mF* is the average F-score of all test images. By binarizing the pixel values of the predicted graph to (0,1), it's able to calculate the precision and recall. The F-score is then calculated as follows.

$$F_\beta = \frac{(1+\beta^2) \times Precision \times Recall}{\beta^2 \times Precision + Recall} \quad (7)$$

$\beta^2$ is set to 0.3 as in [9]. MAE is the average of the absolute error between the predicted map and ground truth. Given a prediction map and the corresponding ground-truth, the algorithm first binarizes both the prediction and the ground truth to [0,1], then the MAE is obtained via (8).

$$MAE = \frac{1}{H \times W} \sum_{i=1}^{H} \sum_{j=1}^{W} |P(i,j) - G(i,j)| \quad (8)$$

In which H and W represent the height and width of the prediction map. $P(i,j)$ and $G(i,j)$ denote the pixel values of the predicted map and ground truth at position $(i,j)$, respectively. S-measure is a recently proposed metric that can evaluate the structural similarity of objects and regions between the predicted map and ground truth. E-measure is a widely used new metric that combines local pixel values with image-level averages to jointly capture image-level statistics and local pixel matching information.

### 4.3. Comparison with State-of-the-arts

In this section, we compare the proposed method with 16 state-of-the-arts of strongly supervised since our approach is a strongly supervised model, including R3Net [33], C2SNet [34], RAS [16], DSS [17], PICA-Net [19], BMPM [15], PAGE [35], TDBU [36], MLMSNet [37], EG-Net [4], PoolNet [38], CPD [39], BASNet [40], F3Net [9], MINet [11] and CSNet [41]. These methods are state-of-the-art algorithms that have been proposed recently, and they share our concerns, such as how to extract and effectively integrate different features with rich semantics.

*4.3.1 Quantitative Comparison:* The comparison between the proposed MPI and the other methods on *mF*, MAE, S-measure, and E-measure is shown in Table 2.

**Table 2** Quantitative comparison of our method with 16 state-of-the-arts over five widely used datasets. The higher *mF*, *S-measure*, *E-measure*, and the lower *MAE* are better, the best results are bolded. For a fair comparison, all results were obtained by testing on saliency maps provided by the authors.

| Datasets<br>Model | ECSSD | | | | PASCAL-S | | | | HKU-IS | | | | DUT-OMRON | | | | DUTS | | | |
|---|---|---|---|---|---|---|---|---|---|---|---|---|---|---|---|---|---|---|---|---|
| | mF | MAE | $S_\alpha$ | $E_\varepsilon$ | mF | MAE | $S_\alpha$ | $E_\varepsilon$ | mF | MAE | $S_\alpha$ | $E_\varepsilon$ | mF | MAE | $S_\alpha$ | $E_\varepsilon$ | mF | MAE | $S_\alpha$ | $E_\varepsilon$ |
| R3Net [33] | .914 | .040 | .910 | .929 | .794 | .100 | .800 | .836 | .893 | .036 | .895 | .939 | .747 | .063 | .815 | .850 | .785 | .057 | .834 | .867 |
| C2SNet [34] | .866 | .054 | .896 | .915 | .763 | .081 | .833 | .840 | .853 | .048 | .888 | .928 | .682 | .071 | .799 | .828 | .718 | .062 | .831 | .846 |
| RAS [16] | .889 | .059 | .893 | .914 | .777 | .104 | .792 | .829 | .874 | .045 | .888 | .931 | .713 | .062 | .814 | .846 | .751 | .059 | .839 | .861 |
| DSS [17] | .873 | .052 | .884 | .908 | .758 | .080 | .797 | .826 | .855 | .039 | .879 | .925 | .673 | .074 | .789 | .819 | .712 | .065 | .826 | .842 |
| PICA-Net [19] | .885 | .044 | .917 | .910 | .789 | .085 | .838 | .828 | .870 | .039 | .908 | .934 | .710 | .065 | .835 | .834 | .749 | .051 | .867 | .852 |
| BMPM [15] | .868 | .044 | .911 | .914 | .762 | .074 | .843 | .841 | .871 | .038 | .907 | .937 | .692 | .064 | .809 | .837 | .745 | .049 | .862 | .860 |
| PAGE [35] | .906 | .042 | .912 | .920 | .806 | .078 | .835 | .841 | .882 | .037 | .903 | .940 | .736 | .066 | .824 | .853 | .777 | .051 | .854 | .869 |
| TDBU [36] | .880 | .040 | .918 | .922 | .775 | .072 | .844 | .850 | .878 | .038 | .907 | .942 | .739 | .059 | .837 | .854 | .767 | .048 | .865 | .879 |
| MLMSNet [37] | .914 | .038 | .911 | .925 | .831 | .069 | .849 | .860 | .892 | .034 | .901 | .945 | .735 | .056 | .817 | .846 | .799 | .045 | .856 | .882 |
| EG-Net [4] | .920 | .041 | .918 | .927 | .817 | .074 | .852 | .848 | .898 | .031 | .918 | .948 | .755 | **.053** | .841 | .867 | .815 | .039 | .875 | .891 |
| PoolNet [38] | .915 | .039 | .921 | .924 | .822 | .074 | .845 | .850 | .899 | .032 | .916 | .949 | .747 | .055 | .835 | .863 | .809 | .040 | .883 | .889 |
| CPD [39] | .917 | .037 | .918 | .925 | .824 | .072 | .842 | .849 | .891 | .034 | .905 | .944 | .747 | .056 | .825 | .866 | .805 | .043 | .869 | .886 |
| F3Net [9] | .925 | .033 | .924 | .927 | .840 | .062 | **.854** | .859 | .910 | .028 | .917 | .953 | **.766** | **.053** | **.838** | **.870** | .840 | **.035** | **.888** | .902 |
| BASNet [40] | .880 | .037 | .916 | .921 | .771 | .076 | .832 | .846 | .895 | .032 | .909 | .946 | .756 | .056 | .836 | .869 | .791 | .047 | .866 | .884 |
| MINet [11] | .924 | .033 | **.925** | .927 | .842 | .064 | .850 | .851 | .908 | .028 | .920 | .953 | .756 | .055 | .833 | .865 | .828 | .037 | .884 | .898 |
| CSNet [41] | .925 | .036 | .924 | .928 | .823 | .068 | .851 | .847 | .902 | .030 | **.921** | .952 | .750 | .055 | **.838** | .861 | .823 | .037 | .884 | .896 |
| Ours | **.933** | **.032** | **.925** | **.930** | **.844** | **.060** | **.854** | **.862** | **.919** | **.026** | **.921** | **.955** | .763 | .055 | .830 | .861 | **.852** | **.035** | **.888** | **.908** |

It can be shown that MPI achieves the best performance on four of the five datasets. Compared with other methods, the proposed model has richer receptive fields, resulting in richer semantics of the generated features. When fusing multiple semantic features, instead of using a top-to-down strategy, such as the HED+U-Net structure, MPI uses a simple parallel fusion approach. This kind of parallel fusion can reduce redundant features from being introduced repeatedly, and since features with similar receptive fields are selected each time, the effect brought by the differences of receptive fields can be weakened when fusing multiple semantic features.

It's noting that the proposed method performs slightly worse on the DUT-OMRON dataset compared with F3Net. We suspect the reason for this is that most of the images in the DUT-OMRON dataset have multiple salient objects, and the image background is very complex. The proposed MRE module enhances the semantic perceptions of the features through dilated convolutions, but the interval sampling of the dilated convolution makes the semantic features lose more details, which makes the model difficult to distinguish the salient object boundaries from the complex background. Fig.7 shows some failure cases of the proposed method on the DUT-OMRON dataset.

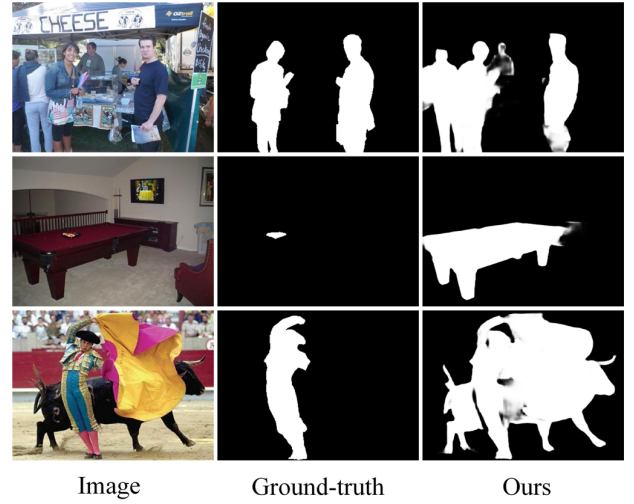

Image      Ground-truth      Ours

***Fig.7.*** *Some failure cases of the proposed method*

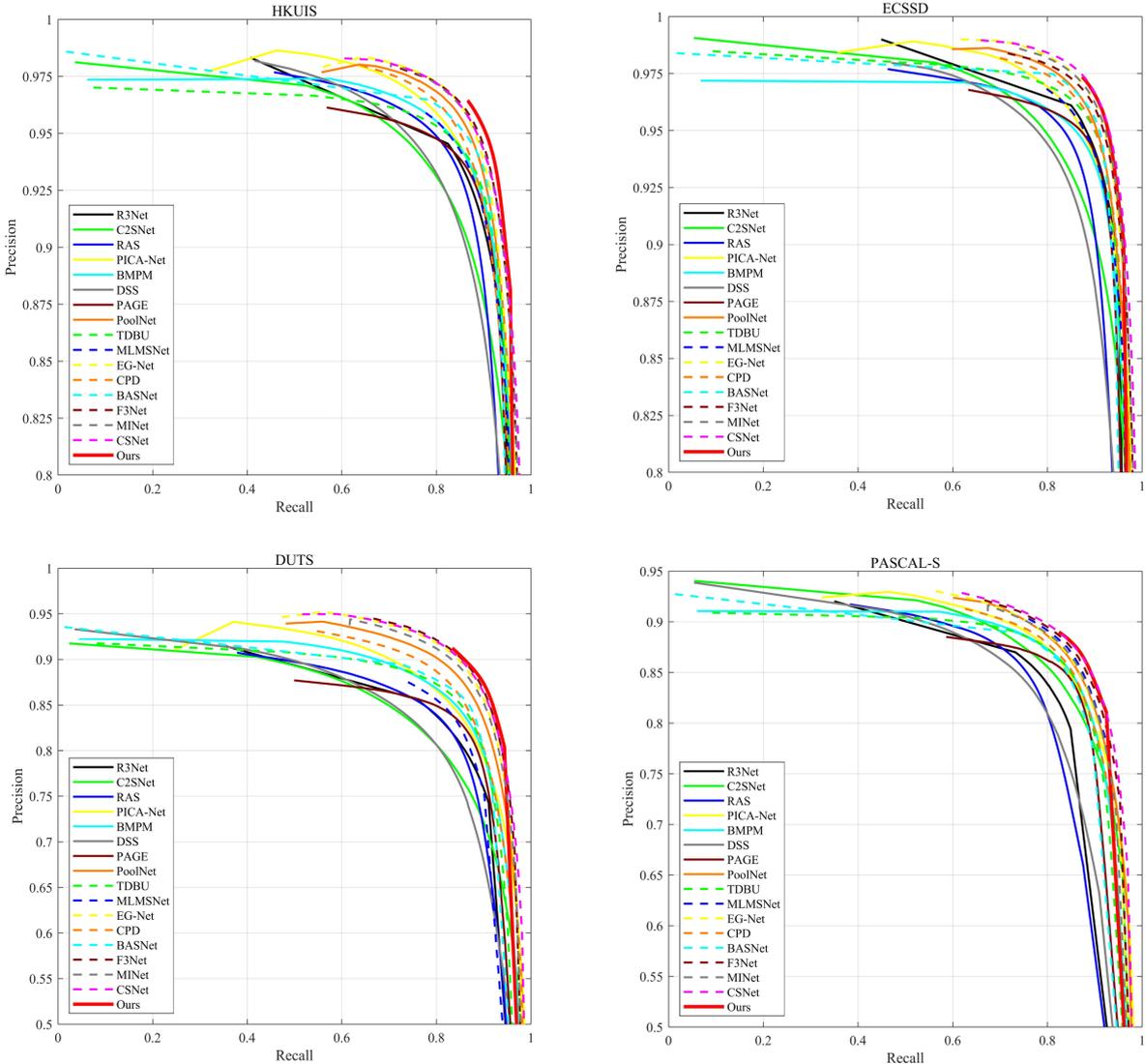

***Fig.8.*** *The precision-recall curves of the proposed method and the 16 comparison methods on PASCAL-S, HKUIS, ECSSD, and DUTS.*

*4.3.2 P-R curve:* We plot the precision-recall curves of the proposed method and the 16 comparison methods on different datasets, as shown in Fig.8. The solid red line represents the proposed method. It can be seen that MPI achieves state-of-the-art performance. Benefit from features with rich receptive fields, as well as parallel fusion strategy of multiple features, the better P-R curves are generated.

*4.3.3 Visual Comparison:* To better verify the superiority of the proposed model and qualitatively analyse the performance of the model, in addition to plotting the P-R curves, we also show some visualization results, as shown in Fig.9. Some challenging test images are selected from different datasets, including images with objects touching boundaries such as the second row, images with multiple objects such as the third, fourth, and eighth rows, and images with low contrast such as the fifth row. It can be seen that our method still performs well on these images, it not only locating the salient objects accurately but also generating clearer saliency maps compared to other methods.

### 4.4. Ablation Studies

This section verifies the gain of multi-receptive fields for semantic features and compares the performance of the proposed PFS with other top-to-down fusion architectures to demonstrate the superiority of the proposed method.

*4.4.1 Multi-receptive features:* The network is trained and validated with and without the MRE module, respectively, holding all other configurations constant. The performance of these two models on the DUTS, PASCAL-S, and HKUIS datasets is shown in Table 3. As seen in Table 3, there are significant differences in performance between the two models on the three datasets. After the addition of MRE, the performance of the model improves significantly, especially on the DUTS dataset, where *mF* is increased by 2.2%. The reason is that MRE increases the receptive fields of features, multi-receptive features can improve the semantic representation and enhance the model's perception of saliency objects. Benefit from the MRE, the model can infer saliency maps with the more accurate object, as shown in Fig 10.

**Table 3** Performance comparison on DUTS, PASCAL-S, and HKUIS dataset.

|  | MRE | *mF* | *MAE* | $S_\alpha$ | $E_\varepsilon$ |
|---|---|---|---|---|---|
| DUTS | × | .830 | .039 | .877 | .895 |
|  | √ | **.852** | **.035** | **.888** | **.908** |
| PASCAL-S | × | .830 | .063 | .847 | .856 |
|  | √ | **.844** | **.060** | **.854** | **.862** |
| HKUIS | × | .913 | .027 | .918 | .952 |
|  | √ | **.919** | **.026** | **.921** | **.955** |

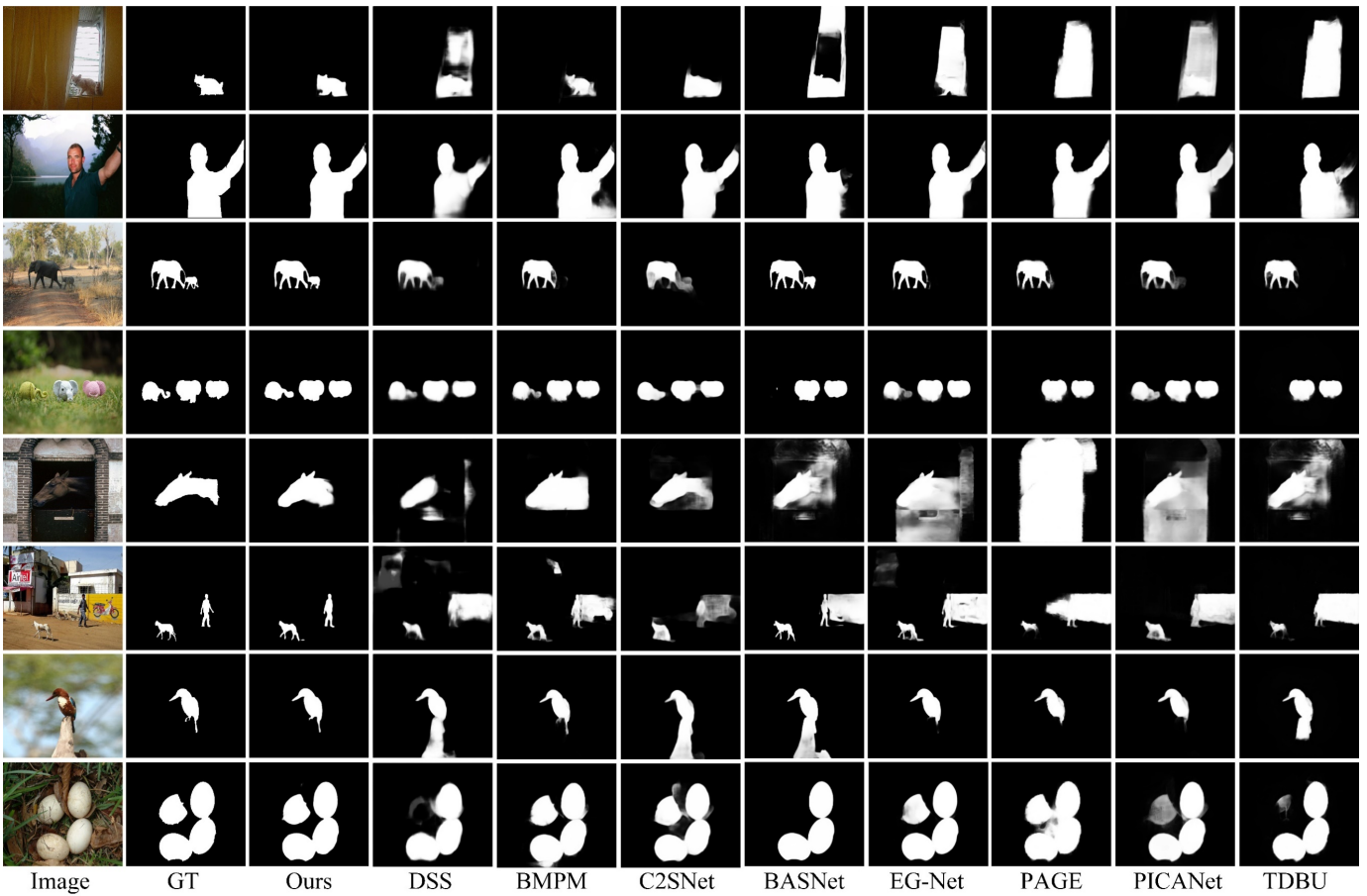

**Fig.9.** *Visualization of comparative results. It can be seen that our method not only accurately locates the salient objects but also generates a clearer saliency map.*

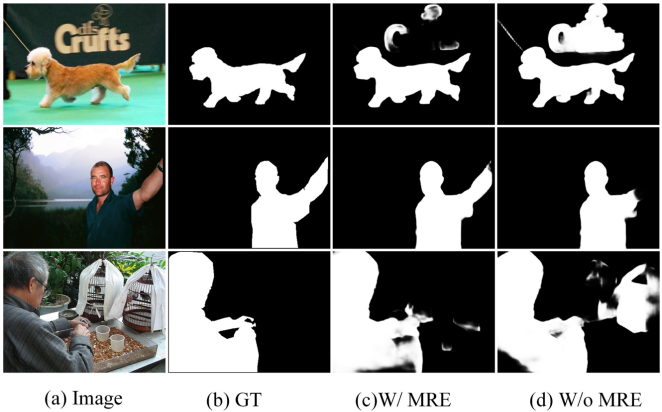

(a) Image  (b) GT  (c) W/ MRE  (d) W/o MRE

***Fig.10.*** *Some saliency maps generated by models with and without MRE, respectively.*

**Table 4** The performance of different fusion architectures on the DUTS dataset.

|  | MRE | $mF$ | $MAE$ | $S_\alpha$ | $E_\varepsilon$ |
|---|---|---|---|---|---|
| U-Net | × | .818 | .041 | .877 | .890 |
|  | √ | .832 | .036 | .886 | .903 |
| HED | × | .808 | .045 | .853 | .888 |
|  | √ | .821 | .041 | .864 | .897 |
| UNet+HED | × | .831 | .040 | .869 | .898 |
|  | √ | .844 | .037 | **.888** | .906 |
| PFS | × | .830 | .039 | .877 | .895 |
|  | √ | **.852** | **.035** | **.888** | **.908** |

*4.4.2 Different feature fusion architectures:* To better evaluate the different architectures, the proposed PFS strategy is compared with the current popular U-Net, HED, and U-Net+HED architectures on DUTS datasets with and without MRE, respectively. The results are shown in Table 4. It can be seen that when MRE is not added, the performance of the proposed PFS is close to that of the U-Net+HED structure and outperforms both the U-Net and HED architectures alone. After adding MRE, the performance of PFS is better than U-net +HED. The reason why PFS outperforms the other three architectures with the addition of MRE is that MRE increases the semantic representation of the features and widens the difference between the features at the same time. Regardless of the architecture, the performance of the model is significantly improved after applying MRE. This demonstrates the effectiveness of MRE and proves that multi-receptive features can enhance the semantic representation and thus the performance of the model.

The result shows that the U-Net structure is superior to the HED architecture under the same conditions, which proves the importance of extracting and fusing multi-scale features for image understanding and salient object detection.

## 5. Conclusion

In this paper, a novel salient object detection method is designed to generate multi-receptive features and fuse multi-scale semantic features. The proposed MRE can efficiently expand the receptive fields of deep features and generate semantic features with multiple receptive fields, which can enhance the semantic representation and improve the model's understanding of the image context. In order to reduce the reuse of redundant information in the complex top-to-down fusion structure and weaken the semantic differences between multi-scale features, this paper designs a relatively simple but effective PFS strategy to fuse features from different layers. Ablation analysis confirms the effectiveness of the proposed modules. Experimental results on five datasets demonstrate the superiority of the proposed method.